\documentclass[a4paper,11pt]{article}

\pdfoutput=1  

\usepackage{graphicx}
\usepackage{url}
\usepackage{times}
\usepackage{natbib}
\usepackage[margin=25mm]{geometry}

\usepackage{tikz}
\usetikzlibrary{positioning,arrows,patterns}
\tikzset{execute at begin node=\strut}
\tikzset{
  center coordinate/.style={
    execute at end picture={
      \path ([rotate around={180:#1}]perpendicular cs: horizontal line through={#1},
                                  vertical line through={(current bounding box.east)})
            ([rotate around={180:#1}]perpendicular cs: horizontal line through={#1},
                                  vertical line through={(current bounding box.west)});}}}

\definecolor{lila}{RGB}{159,114,207}
\definecolor{orange}{RGB}{230,130,50}
\colorlet{plila}{lila!60!white}
\colorlet{porange}{orange!60!white}
\definecolor{braun}{tHsb}{27,1,.5}
\definecolor{violett}{tHsb}{269,1,.5}
\definecolor{hellviolett}{tHsb}{269,1,.8}
\definecolor{grau}{RGB}{192,192,192}

\usepackage{sidecap}
\sidecaptionvpos{figure}{t}
\usepackage{subcaption}
\usepackage{xfrac}
\usepackage{tabularx} 
\newcolumntype{C}{>{\centering\arraybackslash}X} 

\usepackage{hyperref}
\hypersetup{
    colorlinks,
    linkcolor={violett},
    citecolor={braun},
    urlcolor={hellviolett}
}
\usepackage{footnotebackref}
\usepackage[nameinlink,capitalize]{cleveref}
\crefname{equation}{}{}
\crefformat{section}{#2\S#1#3}
\crefformat{subsection}{#2\S#1#3}

\title{Semantic Composition via Probabilistic Model Theory}
\date{}

\author{Guy Emerson and Ann Copestake \\
  Computer Laboratory \\
  University of Cambridge \\
  {\tt \{gete2,aac10\}@cam.ac.uk} \\
}

\begin{document}
\maketitle
\thispagestyle{empty}
\pagestyle{empty}

\begin{abstract}
Semantic composition remains an open problem for vector space models of semantics.
In this paper, we explain how the probabilistic graphical model used in
the framework of Functional Distributional Semantics
can be interpreted as a probabilistic version of model theory.
Building on this, we explain how various semantic phenomena can be recast
in terms of conditional probabilities in the graphical model.
This connection between formal semantics and machine learning is helpful in both directions:
it gives us an explicit mechanism for modelling context-dependent meanings
(a challenge for formal semantics),
and also gives us well-motivated techniques for composing distributed representations
(a challenge for distributional semantics).
We present results on two datasets that go beyond word similarity,
showing how these semantically-motivated techniques
improve on the performance of vector models.
\end{abstract}

\section{Introduction}

Vector space models of semantics are popular in NLP,
as they are easy to work with,
can be trained on unannotated corpora,
and are useful in many tasks.
They can be trained in multiple ways, including count methods \citep{turney2010vector}
and neural embedding methods \citep{mikolov2013vector}.
Furthermore, they allow
a natural and computationally efficient measure of similarity,
in the form of cosine similarity.

However, even if we can train models that produce good similarity scores,
a vector space does not provide natural operations
for other aspects of meaning.
How can vectors be composed
to form semantic representations for larger phrases?
Can we say that one vector implies another?
How do we capture how meanings vary according to context?
An overview of existing approaches to these questions is given in~\cref{sec:rel},
but these issues do not have clear solutions.

In contrast, the framework of Functional Distributional Semantics \citep{emerson2016} (henceforth E\&C)
aims to overcome such issues, not by extending a vector space model,
but by learning a different kind of representation.
Each predicate is represented not by a vector, but by a \textit{function},
which forms part of a probabilistic graphical model.
In~\cref{sec:model}, we build on the description given by E\&C,
and explain how this graphical model
can in fact be viewed as encapsulating a probabilistic version of model theory.
With this connection, we can naturally transfer concepts in formal semantics
to this probabilistic framework,
and we culminate in~\cref{sec:quant}
by showing how generalised quantifiers
can be interpreted in our probabilistic model.
In~\cref{sec:depend}, we look at
how to naturally derive context-dependent representations,
and further, how these representations can be used
for certain kinds of inference and semantic composition.

In~\cref{sec:eval}, we turn to using the model in practice, and evaluate on three tasks.
Firstly, we look at lexical similarity,
to show it is competitive with vector-based models.
Secondly, we consider the dataset produced by \citet{grefenstette2011svo},
which measures the similarity of verbs in the context of a specific subject and object.
Finally, we consider the RELPRON dataset produced by \citet{rimell2016relpron},
which requires matching individual nouns to short phrases including relative clauses.
Our aim is to show that,
not only does the connection with formal semantics give us
well-motivated techniques to tackle these disparate datasets,
but this also leads to improvements in performance.

\pagebreak

\section{Related Work}
\label{sec:rel}

One approach to compositionality in a vector space model
is to find a composition function that maps a pair of vectors
to a new vector in the same space.
\citet{mitchell2010compose} compare a variety of such functions,
but they find that componentwise addition and multiplication are in fact
competitive with the best functions they consider,
despite being symmetric and hence insensitive to syntax.

Another approach is to use a recurrent neural network,
which processes text one token at a time,
updating a hidden state vector at each token.
The final hidden state can be seen as a representation of the whole sequence.
However, the state cannot be directly compared to the word vectors --
indeed, they may have different numbers of dimensions.
Other architectures have been proposed, aiming to use syntactic structure,
such as recursive neural networks \citep{socher2010recursive}.
However, this still does not use semantic structure --
for example, there is no connection between active and passive voice sentences.

\citet{coecke2010tensor} and \citet{baroni2014tensor} introduce a tensor-based approach,
where words are represented not just by vectors,
but also by higher-order tensors, which combine according to argument structure:
nouns are vectors,
intransitive verbs are matrices (mapping noun vectors to sentence vectors),
transitive verbs are third-order tensors (mapping pairs of noun vectors to sentence vectors),
and so on.
However, \citet{grefenstette2013tensor} showed that
quantifiers cannot be expressed in this framework.

Furthermore, in all the above methods, it is unclear how to perform inference,
While we can use the representations as input features for another system,
they do not have an inherent logical interpretation.
\citet{balkir2016sentence} extend the tensor-based framework to allow inference,
but rely on existing vectors, and must assume the dimensions have logical interpretations.
\citet{lewis2013logic} use distributional information to cluster predicates,
but this leaves no graded notion of similarity.
\citet{garrette2011logic} and \citet{beltagy2016logic}
incorporate a vector space model into a Markov Logic Network,
in the form of weighted inference rules
(the truth of one predicate implying the truth of another).
However, this assumes we can interpret similarity in terms of inference
(a position defended by \citet{erk2016alligator}),
and requires existing vectors,
rather than directly learning logical representations from distributional data.

Many proposals exist for contextualising vectors.
\citet{erk2008context} and \citet{thater2011context}
modify a vector according to syntactic dependencies.
However, by proposing new operations,
they make assumptions about the properties of the space,
which may not apply to all models.
\citet{erk2010exemplar} build a context-specific vector,
by combining the most similar contexts in a corpus.
However, this reduces the amount of training data.
\citet{lui2012usage}'s ``per-lemma'' model uses Latent Dirichlet Allocation
to model contextual meaning as a mixture of senses,
but this requires training a separate LDA model for each word.
Furthermore, all of these methods focus on a specific kind of context,
making it nontrivial to generalise them to arbitrary contexts.

Our notion of probabilistic truth values
is similar to the Austinian truth values in
the framework of probabilistic Type Theory with Records (TTR) \citep{cooper2005type,cooper2015prob}.
\citet{sutton2015prob,sutton2017prob} takes a similar probabilistic approach to truth values
to deal with philosophical problems concerning gradable predicates.
Our stochastic generation of situations
is also similar to the approach taken by \citet{goodman2015prob},
who represent semantics with the stochastic lambda calculus,
using hand-written probabilistic models to show how semantics and world knowledge can interact.
While these approaches are in principle compatible with our work,
they do not provide an approach to distributional semantics.
We use Minimal Recursion Semantics \citep{copestake2005mrs},
as it can be represented using dependency graphs --
this allows a more natural connection with probabilistic graphical models,
as explained in~\cref{sec:sem-func}.

Others have also proposed representing the meaning of a predicate as a classifier.
\citet{larsson2013classifier} represents the meaning of a perceptual concept
as a classifier of perceptual input, in the TTR framework.
\citet{schlangen2016classifier} train image classifiers using captioned images,
and \citet{zarriess2017classifier,zarriess2017classifier2} build on this,
using distributional similarity to help train such classifiers.
However, they do not learn an interpretable representation directly from text;
rather, they use similarity scores
to generalise from one label of an image to other similar labels.
\citet{mcmahan2015colour} represent the meaning of a colour term
as a probabilistic region of colour space,
which could also be interpreted as a probabilistic classifier.
However, this model was not intended to be a general-purpose distributional model.

\pagebreak

\section{From Model Theory to Probability Theory}
\label{sec:model}

In this section, we show how model theory can be recast in a probabilistic setting.
The aim is not to detail a full probabilistic logic,
but rather to show how we can define a family of probability distributions
that capture traditional model structures as a special case,
while also allowing structured representations of the kind used in machine learning.
In this way, we will be able to view Functional Distributional Semantics
as a generalisation of model theory.

\vspace*{-1mm}
\subsection{Background: Model Theory, Neo-Davidsonian Events, and Situations}
\label{sec:back}

A standard approach to formal semantics
is to use an extensional model structure
\citep{cann1993sem, allan2001sem, kamp2013sem}.
We first define a set of `individuals' (or `entities') in the model.
We then define the meaning of a predicate to be its extension --
the subset of individuals for which the predicate is true.
The extension can also be characterised in terms of a truth-conditional function --
a function mapping from individuals to truth-values.
Individuals in the extension of the predicate are mapped to true,
and all other individuals to false.


We take a neo-Davidsonian approach to event semantics
\citep{davidson1967event, parsons1990event}.
This treats events as also being individuals,
and verbal predicates are one-place relations,
which can be true of event individuals.
Other participants in an an event are indicated by two-place relations,
linking the event to the argument individual.
For example, a sentence like \textit{pictures tell stories}
would be represented with three individuals and five relations:
$\textit{picture}(x)$,
$\textit{tell}(y)$,
$\textit{story}(z)$,
$\textsc{arg1}(y,x)$,
$\textsc{arg2}(y,z)$.
Here, the \textsc{arg1} and \textsc{arg2} relations
express the argument structure of the telling event.

Finally, we take an approach in the spirit of situation semantics \citep{barwise1983situation},
and assume that the model contains a set of \textit{situations}.
Each situation consists of a small number of individuals
(unlike a possible world, which would consist of many individuals),
and the relations that stand between them.
As we are taking a neo-Davidsonian approach,
this means that we take a situation to be set of individuals,
where each predicate assigns a truth value to each individual,
and where there are two-place relations between individuals,
to express argument structure.



\vspace*{-1mm}
\subsection{Model Structures as Probability Distributions}
\label{sec:model-prob}

In this section, we generalise this notion of a model structure in two ways.
Firstly, rather than a set of situations,
we will consider a probability distribution over a set of situations.
Secondly, rather than deterministic truth-conditional functions,
we will consider probabilistic truth-conditional functions.

A probability distribution over a set of situations
is naturally more general than the set itself,
since it provides more information --
as well as knowing that a situation is in the set,
we additionally know its probability.
From the formal linguistic point of view, this might seem irrelevant to the notion of truth.
However, from the machine learning point of view
(and perhaps also from the acquisition point of view),
it is very helpful --
if our aim is not just to \textit{represent} what is true,
but also to \textit{learn} what is true,
we do not know in advance what situations should be part of the model structure.
By using probability distributions,
we can smoothly change between different models.
This lets us use continuous optimisation algorithms,
such as methods based on gradient descent,
which are generally more efficient than discrete optimisation algorithms.
Intuitively, as we learn about what kinds of situations exist,
we can update the model appropriately.

A truth-conditional function can be defined as
a function mapping from a set of individuals to the set $\{0,1\}$,
where $0$ denotes falsehood, and $1$ truth.
We can generalise this to
a function mapping from a set of individuals to the range $[0,1]$.
This allows us to naturally model the fuzzy boundaries of concepts,
by using intermediate values between $0$ and $1$.
This idea was used by \citet{labov1973cup}
to model the fuzzy boundaries between concepts like \textit{cup}, \textit{mug}, and \textit{bowl}.
For an unusual object that is intermediate between
a typical cup and a typical bowl,
we can say that the predicates for \textit{cup} and \textit{mug}
both have an intermediate probability of being true of the object.
As with our previous generalisation step,
allowing a continuous range of values is also helpful during learning,
since we can smoothly change a function between assigning truth or falsehood to a particular individual.

\subsection{Denotations versus Truth-Conditional Functions}
\label{sec:truth}

If individuals are atomic elements, without any further structure,
then denotations and truth-conditional functions have almost identical representations.
A denotation is a subset of the set of individuals,
while a truth-conditional function is the indicator function for this subset:
individuals in the denotation are mapped to $1$, and other individuals to $0$.
Converting between these two representations is trivial.

However, if individuals are structured objects,
denotations and truth-conditional functions may have rather different representations.
To represent the structure of individuals, we assume we have a semantic space,
where each point in the space represents a possible individual,
including information about all its features.
We will use the term `pixie' to refer to a point in the semantic space,
as it is intuitively a `pixel' of the space.
Note that E\&C use the term `entity' to refer to both individuals and pixies.

For example, consider a model with five individuals:
two black cats, a white cat, a bowl of rice, and a carrot.
If we use a semantic space, and represent these individuals with the features
\textsc{colour} (\textit{black}, \textit{white}, or \textit{orange})
and \textsc{animacy} ($+$ or $-$),
then the denotation of the predicate for \textit{cat} is a set of three individuals,
whose pixies are
\{\textsc{colour}:~\textit{black}, \textsc{animacy}:~$+$\} (appearing twice) and
\{\textsc{colour}:~\textit{white}, \textsc{animacy}:~$+$\} (appearing once).\footnote{%
  We could add an `ID' feature to distinguish otherwise identical individuals,
  but will not take this approach here.
}
As a probability distribution,
the denotation assigns a probability of \sfrac{2}{3} to the first pixie,
\sfrac{1}{3} to the second,
and $0$ to all others.
However, the truth-conditional function can be represented much more simply --
it takes the value $1$ if and only if the pixie is animate.

As can be seen in this example,
we may have multiple individuals represented by the same pixie.
This can be accounted for in the probabilistic model structure,
by assigning higher probabilities to pixies that correspond to more individuals.
Note that, technically, this means that we are not working directly with distributions over situations,
but rather with distributions over equivalence classes of situations,
where situations are equivalent if their individuals are indistinguishable in terms of their features.


\subsection{Functional Distributional Semantics as Model-Theoretic Semantics}
\label{sec:sem-func}

\begin{SCfigure}[50]
\centering

\begin{tikzpicture}[on grid, node distance=21mm and 25mm, x=25mm, y=21mm, style=thick]
\tikzstyle{word}=[inner sep=1mm]

\draw[color=white] (-0.5, 0) rectangle (2.5, 0) ;

\node[word] (x) {picture};
\node[word, right=of x] (y) {tell};
\node[word, right=of y] (z) {story};

\draw[-angle 60] (y) -- (x) node[midway, above, xshift=1mm] {\textsc{arg1}} ;
\draw[-angle 60] (y) -- (z) node[midway, above, xshift=-1mm] {\textsc{arg2}} ;

\end{tikzpicture}

\caption{A simplified DMRS graph,
which could be generated by \cref{fig:graph} below.
Such graphs are observed during training.
}
\label{fig:dmrs}

\end{SCfigure}
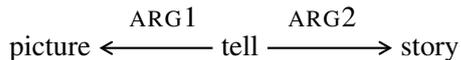

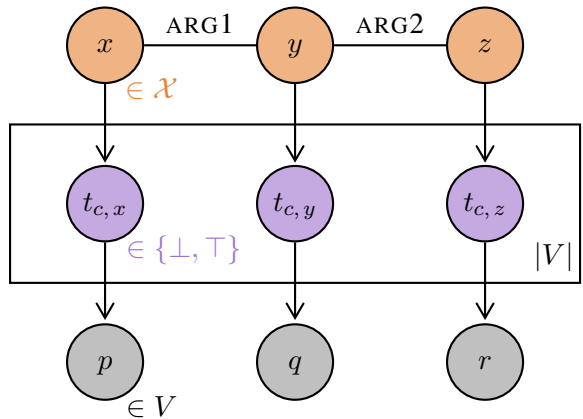
\begin{SCfigure}[50]
\centering

\begin{tikzpicture}[on grid, node distance=21mm and 25mm, x=25mm, y=21mm, style=thick, inner sep=0pt]
\tikzstyle{ent}=[circle, draw, minimum size=10mm, fill=porange]
\tikzstyle{val}=[circle, draw, minimum size=10mm, fill=plila]
\tikzstyle{tok}=[circle, draw, minimum size=10mm, fill=grau]


\node[ent] (y) {$y$} ;
\node[ent, right=of y] (z) {$z$} ;
\node[ent, left=of y] (x) {$x$} ;

\draw (y) -- (z) node[midway, above, color=black] {\textsc{arg2}} ;
\draw (y) -- (x) node[midway, above, color=black] {\textsc{arg1}} ;

\node[below right=3.5mm and 2.5mm of x, anchor=north west] {\textcolor{orange}{$\in \mathcal{X}$}} ;


\draw (-1.5,-0.5) rectangle (1.5,-1.5) ;

\node[val, below=of x] (tx) {$t_{c,\,x}$} ;
\node[val, below=of y] (ty) {$t_{c,\,y}$} ;
\node[val, below=of z] (tz) {$t_{c,\,z}$} ;

\tikzset{style=ultra thick}
\draw[-angle 60] (x) -- (tx);
\draw[-angle 60] (y) -- (ty);
\draw[-angle 60] (z) -- (tz);

\node[below right=3.5mm and 2.5mm of tx, anchor=north west] {\textcolor{lila}{$\in \left\{\bot,\top\right\}$}} ;
\node[xshift=-2ex, yshift=2ex] at (1.5, -1.5) {$|V|$};


\node[tok, below=of tx] (a) {$p$} ;
\node[tok, below=of ty] (b) {$q$} ;
\node[tok, below=of tz] (c) {$r$} ;

\draw[-angle 60] (tx) -- (a);
\draw[-angle 60] (ty) -- (b);
\draw[-angle 60] (tz) -- (c);

\node[below right=3.5mm and 2.5mm of a, anchor=north west] {$\in V$} ;

\end{tikzpicture}

\caption{
Probabilistic graphical model for Functional Distributional Semantics
(E\&C, Fig.~3).
Each node denotes a random variable.
The plate (box in middle row) denotes repetition of random variables. \newline
\textbf{Top row:} pixies $x$, $y$, and $z$, randomly drawn from a semantic space~$\mathcal{X}$.
Their joint distribution is determined by the DMRS links. \newline
\textbf{Middle row:} each predicate $c$ in the vocabulary $V$ is randomly
true or false for each pixie,
according to the predicate's semantic function. \newline
\textbf{Bottom row:} for each pixie, we randomly generate one predicate,
out of all predicates true of the pixie.
\vspace*{-8mm}
}
\label{fig:graph}

\end{SCfigure}

Now we have described the above probabilistic generalisation of a model structure,
we explain how Functional Distributional Semantics can be seen as implementing such a generalised model structure.

E\&C define a probabilistic graphical model
to generate semantic dependency graphs like that in \cref{fig:dmrs}.
The aim is to train the model in an unsupervised\footnote{%
  Following \citet{ghahramani2004unsupervised},
  supervised learning requires both inputs and outputs, while
  unsupervised learning requires only inputs.
  The annotations in our training corpus are not desired outputs,
  so learning is unsupervised in this sense.
}
way on a parsed corpus --
that is, to optimise the model parameters
to maximise the probability of generating the dependency graphs in the corpus.
Furthermore, Dependency Minimal Recursion Semantics
(DMRS) \citep{copestake2009dmrs}
allows a logical interpretation of the dependency graphs:
each node represents a predicate,
and the \textsc{arg} links represent argument structure.
The graphical model in \cref{fig:graph}
generates dependency graphs corresponding to transitive sentences --
the predicates ($p$, $q$, $r$) can be seen at the bottom,
and the dependency links (\textsc{arg1}, \textsc{arg2}) can be seen at the top.
For example, $p$, $q$, and $r$ might correspond to
\textit{pictures}, \textit{tell}, and \textit{stories}.

Rather than generating a dependency graph directly,
the semantic function model assumes that it generated based on latent structure.
We assume that for each observed predicate,
there is an unobserved, latent pixie which the predicate is true of.
These pixies are the orange nodes at the top of \cref{fig:graph}.
Each pixie node is a random variable,
taking values in the semantic space~$\mathcal{X}$ of all possible pixies.
We also assume that every predicate is either true or false of each pixie.
These truth values are the purple nodes in the middle row of \cref{fig:graph}
(note that each node is repeated $|V|$ times, once for each predicate).
They are random variables, with two possible values: true or false.
While we know each observed predicate is true of its pixie,
the truth values for all other predicates are latent variables.

The generative model proceeds from the top to the bottom of \cref{fig:graph}.
First, we define a joint distribution over pixies,
as an undirected graphical model --
whenever a pair of pixie nodes is linked,
the model determines how likely it is for specific values of those nodes to co-occur.
This allows us to generate tuples of pixies.
E\&C implement this with a Cardinality Restricted Boltzmann Machine (CaRBM) \citep{swersky2012carbm}:
pixies are sparse binary-valued vectors,
with each dimension representing a different feature.
Each dependency link determines how likely it is for specific features of the linked pixies to co-occur;
this is encoded using one trainable parameter for each pair of dimensions.

Next, we define a semantic function for each predicate --
this maps each pixie to the probability that the predicate is true of it.
So, given a set of generated pixies,
we can generate truth values for each pixie.
E\&C implement these functions with one-layer feedforward networks --
by using a sigmoid activation, the output is in the range $[0,1]$,
so it can be interpreted as a probability.
Finally, given the truth values for all predicates,
we generate one predicate for each pixie,
by choosing from the true predicates.


The above generative process was given by E\&C.
However, we can see the first two stages in this process
as an instance of the probabilistic model structure discussed in~\cref{sec:model-prob}.
The linked pixies can together be viewed as a situation.
The joint distribution over pixies then gives us a distribution over situations,
which can be seen as our probabilistic generalisation of a set of situations in a model structure.
Furthermore, as the semantic functions map from pixies to probabilities,
they can be seen as generalised truth-conditional functions.
So, we can view the semantic function model
as generating dependency graphs based on a probabilistic model structure.
In this model, a denotation can be represented by
a probability distribution over the semantic space,
while a truth-conditional function can be represented by a semantic function,
mapping from the semantic space to~$[0,1]$.


However, we should note that this model only implements soft constraints on semantics --
indeed, it would be difficult to learn hard constraints from corpus data alone.
This means that, all our distributions over pixies
have a non-zero probability for every pixie,
and all our semantic functions assign a non-zero probability of truth to every pixie.
By analogy with a traditional model structure,
we might want to have zero values, to indicate that a certain pixie or situation is impossible,
or that a certain predicate is definitely false.
However, from a Bayesian point of view, zero probabilities are problematic --
they would imply that, no matter what new evidence we observe,
we cannot change our mind.

In practice, some probabilities will be vanishingly small.
In fact, to make interesting predictions, this is necessary --
for high-dimensional spaces, an interesting subspace
(perhaps representing a domain, like rock-climbing or ballroom dancing)
may be small.
For example, suppose we have 1000 binary-valued dimensions, with only 40 active at once.
This gives $10^{72}$ pixies.
A subspace only using 200 dimensions has $10^{42}$ pixies,
or one part in $10^{30}$ of the whole space!
To define a distribution with most probability mass in this subspace,
pixies in the subspace must be at least $10^{30}$ times more likely than outside.

Suppose a predicate is probably true in this subspace,
and probably false outside.
Given a uniform prior over the space,
and observing the predicate to be true,
we may expect the posterior to assign most probability mass to the subspace.
For this to happen, the probability of truth in the subspace
must be $10^{30}$ times larger than outside.
So, for a semantic function to be useful,
it must be close to a step function.
This makes it look more like
a traditional truth-conditional function with only $0$ and $1$ as values.


\pagebreak

\subsection{Interpretation of Quantifiers}
\label{sec:quant}
\vspace*{-1mm}

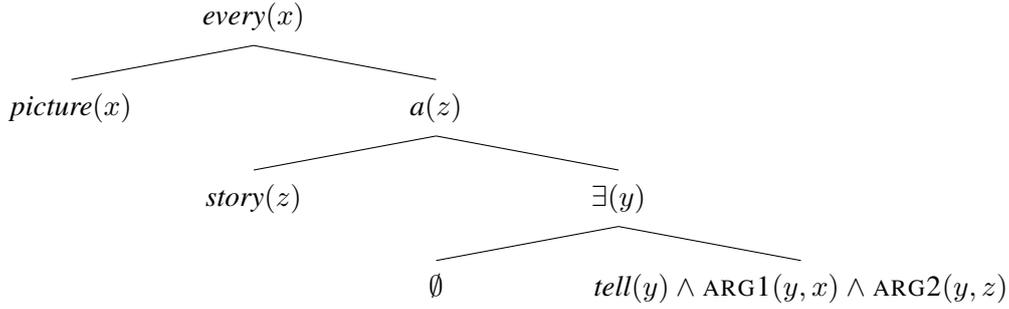
\begin{figure}
\center
\begin{tikzpicture}[on grid, node distance = 12mm and 24mm]
\node (every) {$\textit{every}(x)$} ;
\node [below left = of every] (picture) {$\textit{picture}(x)$} ;
\node [below right = of every] (a) {$\textit{a}(z)$} ;
\node [below left = of a] (story) {$\textit{story}(z)$} ;
\node [below right = of a] (exists) {$\exists(y)$} ;
\node [below left = of exists] (empty) {$\emptyset$} ;
\node [below right = of exists] (tell) {$\textit{tell}(y)\wedge\textsc{arg1}(y,x)\wedge\textsc{arg2}(y,z)$} ;
\draw (every.south) -- (picture.north) ;
\draw (every.south) -- (a.north) ;
\draw (a.south) -- (story.north) ;
\draw (a.south) -- (exists.north) ;
\draw (exists.south) -- (empty.north) ;
\draw (exists.south) -- (tell.north) ;
\end{tikzpicture}
\vspace*{-1mm}
\caption{
Fully scoped representation of the most likely reading of \textit{Every picture tells a story}.
Each non-terminal node is a quantifier, its left child its restriction, and its right child its body.
We assume that event variables are existentially quantified with no constraints on the restriction.
}
\label{fig:scope}
\vspace*{-4mm}
\end{figure}

Interpreting the semantic function model as a probabilistic model structure,
we can define quantification in a natural way.
Unlike \citet{herbelot2015quantifier},
we are not mapping from a distributional space to a model structure,
but directly interpreting quantifiers in our distributional model.

To assign a truth value to DMRS graph,
we must first convert it to a fully scoped representation,
such as in \cref{fig:scope}.
In cases of scope ambiguity,
a single DMRS graph allows several scoped representations,
and this conversion must resolve the ambiguity.\footnote{%
	A DMRS graph includes scopal constraints,
	specifying that nodes are the same place in the scope tree,
	or that one dominates another.
	These constraints were not used in the generative model in~\cref{sec:sem-func},
	akin to other simplified MRS-based dependency structures such as EDS \citep{oepen2006eds},
	but they are necessary to specify the correct set of scope readings.
}

In a complete structure, there is one quantifier for each MRS variable,
and hence also for each pixie-valued random variable, as there is a one-to-one mapping between them.
For each quantifier, we define a binary-valued random variable,
representing whether the quantified expression is true (given any remaining free variables).
We define distributions for these random variables recursively,
bottom-up through the scope tree.
At each stage, we marginalise out the quantified pixie variable.
This is analogous to semantic composition in traditional models --
truth values are calculated bottom-up,
and evaluating a quantifier removes a free variable.
At the root of the tree, we have a single probabilistic truth value.

Each quantifier depends on its restriction and body,
each of which may be either\footnote{%
	More generally, we may have a set of predicates and quantified expressions.
	In this case, we can condition on all truth values in the set.
	We consider a single random truth value, for ease of exposition.
}
a predicate or a quantified expression --
both are binary-valued random variables.
In the classical theory of generalised quantifiers,
the truth of a quantified expression depends on
the cardinality of the restriction set, and
the cardinality of the intersection of the restriction and body sets
\citep{barwise1981quant,vanbenthem1984quant}.
As explained in~\cref{sec:truth},
our probabilistic model structure uses probabilities in place of cardinalities.
Let the probabilistic truth values for the quantified expression, restriction, and body
be $Q$, $R$, and $B$, respectively.
It makes sense to consider the conditional probability $P(B|R)$,
which naturally uses both of the classical sets, since
${P(B|R) = \frac{P(R\,\cap B)}{P(R)}}$.
Intuitively, the truth of $Q$ depends on how likely $B$ is to be true, given that $R$ is true.\footnote{%
	This account does not cover cardinal quantifiers.
	However, the English Resource Grammar (ERG) represents numbers not as quantifiers, but as additional predicates.
	This is compatible with \citet{link2002plural}'s lattice-theoretic approach,
	which allows reference to plural individuals without quantification.
	For more information on the semantic analyses in the ERG,
	see the documentation produced by \citet{flickinger2014ers},
	which is available here: \url{http://www.delph-in.net/esd}
}

In a traditional logic, the truth value of a quantified expression
is a function of all free variables.
Analogously, each quantifier's random variable is conditionally dependent on all free variables.
More precisely, if the set of free variables is $V$,
let us define the function $q(V)=P(B|R,V)$.
We can now consider different quantifiers,
and define the probability of the quantified expression $Q$ being true,
in terms of the value of $q$:
\textit{every} is true iff $q=1$,
\textit{some} is true iff $q>0$,
\textit{most} is true iff $q>\sfrac{1}{2}$, and so on.
In these cases, the probability of truth is exactly $0$ or $1$,
but by using intermediate probabilities, we can also naturally model `fuzzy' quantifiers
such as \textit{few} and \textit{many},
which do not have a sharp cutoff.

\pagebreak

\section{From Conditional Dependence to Context Dependence}
\label{sec:depend}

In the previous section, we saw how a model structure can be generalised using probability distributions.
In this section, we show how this approach allows us to capture context-dependent meanings
using conditional probabilities, in a natural way.

\subsection{Occasion Meaning versus Standing Meaning}
\label{sec:context}

When discussing context dependence
(a challenge for both formal semantics and vector-based semantics),
it is helpful to distinguish two kinds of meaning, following \citet{quine1960meaning}:
\textit{standing} meaning refers to the fixed, unchanging meaning that a linguistic expression has in general;
\textit{occasion} meaning refers to the particular meaning that a linguistic expression has in a given context.

\citet{searle1980meaning} discusses an interesting set of examples,
noting how a gardener cutting grass
involves a very different kind of cutting from a child cutting a cake.
There is something common to both events,
but they involve different tools and different physical motions.
However, \citeauthor{searle1980meaning} also notes
how there are also less obvious interpretations of these expressions.
For a gardener who sells turf to people who need ready-grown lawns,
cutting grass could also refer to cutting out an area of grass, including the soil.\footnote{%
	There are yet other interpretations of \textit{cut grass},
	such as \textit{adulterate marijuana},
	but we focus on the two discussed by \citeauthor{searle1980meaning}.
}
This kind of cutting would more closely resemble cutting a cake.
We can see from this example that
while an expression may refer to quite different situations,
some situations may be more likely than others.

This state of affairs can be modelled in our probabilistic version of model theory.
We can say that an expression like \textit{cut grass} has fixed truth conditions --
it is true of both mowing a lawn and preparing a section of turf.
However, we can also say that the former is much more probable than the latter.
More precisely, our prior distribution over situations assigns
a much higher probability to lawn-mowing situations
than to turf-slicing situations;
but the probabilistic truth-conditional functions for \textit{cut} and \textit{grass}
assign high probabilities to their respective individuals in both of these types of situation.

While the expression \textit{cut grass} could refer to different types of situation,
in most contexts we can infer that
it is likely to refer to a lawn-mowing situation.
We can view this as performing Bayesian inference.
We begin with a prior probability distribution over situations,
where the situation includes (at least) two pixies $y$ and $z$,
with an \textsc{arg2} link from $y$ to $z$
(i.e.\ the situations with enough structure for \textit{cut grass},
since \textit{grass} is the \textsc{arg2} of \textit{cut}).
This prior distribution is given to us by our probabilistic model structure.
Importantly, these situations are jointly distributed with
truth values for the predicates for \textit{cut} and \textit{grass} (as well as all other predicates).
On observing that the \textit{cut} predicate is true of the pixie $y$,
and the \textit{grass} predicate is true of the pixie $z$ (the \textsc{arg2} of $y$),
we can form a posterior distribution over situations
(i.e.\ a joint posterior over the pixies).
This posterior should assign a high probability to lawn-mowing situations,
a low probability to turf-slicing situations,
and an extremely low probability to unrelated situations like baking a cake.
If we receive information that makes a turf-slicing situation more likely
(e.g.\ hearing about a person selling turf),
we can update our posterior again, and assign a higher probability to such a situation.
However, in the absence of such information,
we effectively `default' to the higher-probability lawn-mowing interpretation.

In summary, we can view truth-conditional functions as
representing context-invariant standing meanings,
and posterior distributions over situations as
representing context-dependent occasion meanings.

\subsection{Context Dependence in Functional Distributional Semantics}
\label{sec:meanfield}

In Functional Distributional Semantics,
the standing meaning of a predicate is its semantic function --
a mapping from the semantic space to probabilities of truth.
These are the arrows in \cref{fig:graph}
from the orange pixie nodes (top row) to the purple truth value nodes (middle row).
These functions are implemented as feedforward neural networks --
note that the meaning is not the input or output of a network, but rather the network itself.

The occasion meaning of a predicate
is the posterior distribution over the semantic space,
for the pixie the predicate is true of.
The pixies are the orange nodes in \cref{fig:graph},
but these nodes do not directly stand for meanings --
an occasion meaning is the posterior distribution of such a node,
when conditioned on the truth of one or more predicates.

We should also note that,
while we only consider specific kinds of linguistic contexts in this paper,
this approach generalises to arbitrary contexts.
As an occasion meaning is simply a posterior distribution,
we could in principle condition on any kind of observation.
For example, if we are dealing with a specific domain,
and we know the kinds of pixies that are likely to appear in this domain,
we can produce a domain-specific meaning,
which we could then further condition on a linguistic context.

To calculate occasion meanings,
we need to calculate posterior distributions over the semantic space,
given some observed truth values.
However, exactly calculating the posterior is generally intractable,
as this requires summing over the entire semantic space.
For a large number of dimensions, the space is simply too big.
Sampling from the space using a Markov Chain Monte Carlo method,
as described by E\&C,
is also computationally expensive.

To make calculating the posterior tractable,
we can use a variational approximation --
this involves specifying a restricted class of distributions which is easier to work with,
and then finding the optimal distribution in this restricted class
that approximates the posterior.
In particular, we can use a mean field approximation,
following \citet{emerson2017} (henceforth E\&C2) --
we assume that each dimension (intuitively, each feature)
has an independent probability of being active,
and we optimise each of these probabilities based on the mean activations of all other dimensions.
Under this approximation, an occasion meaning is represented by a mean field vector.
Intuitively, we assign high probabilities to a dimension for two possible reasons:
either it's connected with high weights to highly probable dimensions in other pixies,
or activating this dimension makes it much more likely for an observed predicate to be true.
If neither of these facts hold, we will assign a low probability --
because we are enforcing sparsity on the pixie vectors,
the dimensions are effectively competing with each other.

This mean field approximation
gives us a tractable way to approximately calculate a posterior distribution over pixies.
This allows us to construct vectors representing context-dependent meanings,
which we can use as the basis for further calculations,
as illustrated in the following sections.

\subsection{Semantic Composition using Context-Dependent Meanings}
\label{sec:comp}

Semantic composition involves taking semantic representations for multiple expressions,
and combining them into a single representation for the whole expression.
In vector space models, this involves mapping two or more vectors to a single vector.
Intuitively, with a fixed number of dimensions, this loses information.
As \citet{mooney2014cram} colourfully put it,
``You can't cram the meaning of a whole  
\%\&!\$\# sentence into a single \$\&!\#* vector!''
More precisely, if nearby vectors represent similar meanings,
only the first few significant digits of each dimension are important,
which limits how much information a vector can contain.
Even if we use sparse vectors,
at some point we will have `used up' all of the available dimensions.
So, composing vectors is not viable in the general case --
even proponents of vector space models
would not suggest composing vectors
to produce an accurate representation of an entire book.

Unlike vectors, the representations used in formal semantics are not bounded in size --
logical formulae and semantic dependency graphs can be arbitrarily large.
While it can be useful to summarise a large representation with a smaller one,
we do not believe that a semantic theory should force composition to involve summarisation.
Full and detailed semantic representations should also have their place.

In a semantic function model,
we can use DMRS composition.
Individual lexical items are associated with predicates,
and these are composed to form a DMRS graph.
However, the probabilistic framework gives us a new interpretation of the DMRS graphs.
If we start from two DMRS graphs,
we can consider the two posterior distributions over situations defined by those graphs.
Once we compose these two graphs,
we have a new posterior distribution, over larger situations.
However, this posterior is not the same as naively combining the posteriors of the two subgraphs.
As the pixie nodes of the two subgraphs are now linked together,
we have a joint distribution for all the pixie nodes,
which depends on all the observed predicates.
This means that, as we build a composed DMRS graph,
we modify the posterior distributions at every step.
In this way, we can see semantic composition
as simultaneously composing the logical structure
and refining the context-dependent meanings.


\pagebreak

\subsection{Inference using Context-Dependent Meanings}
\label{sec:infer}

\begin{figure}
\centering

\begin{subfigure}[t]{0.49\textwidth}
\centering
\begin{tikzpicture}[on grid, node distance=21mm and 25mm, x=25mm, y=21mm, style=thick, inner sep=0pt,
center coordinate = (x)]
\tikzstyle{ent}=[circle, draw, minimum size=10mm, fill=porange]
\tikzstyle{val}=[circle, draw, minimum size=10mm, fill=plila]
\tikzstyle{tok}=[circle, draw, minimum size=10mm, fill=grau]


\node[ent] (x) {$x$} ;

\node[below right=3.5mm and 2.5mm of x, anchor=north west] {\textcolor{orange}{$\in \mathcal{X}$}} ;


\node[val, below=of x] (tx) {$t_{p,\,x}$} ;

\tikzset{style=ultra thick}
\draw[-angle 60] (x) -- (tx);

\node[below right=3.5mm and 2.5mm of tx, anchor=north west] {\textcolor{lila}{$\in \left\{\bot,\top\right\}$}} ;


\node[val, above=of x] (new) {$t_{a,\,x}$} ;
\draw[-angle 60] (x) -- (new);

\node[below right=3.5mm and 2.5mm of new, anchor=north west] {\textcolor{lila}{$\in \left\{\bot,\top\right\}$}} ;

\node[right = 9mm of new] {\LARGE?};

\end{tikzpicture}

\caption{
Inferring if $a$ is true of $x$,
when there are no further pixies in the situation,
and given the truth value for the predicate $p$.
}
\label{fig:infer-one}
\end{subfigure}
\hfill
\begin{subfigure}[t]{0.49\textwidth}
\centering
\begin{tikzpicture}[on grid, node distance=21mm and 25mm, x=25mm, y=21mm, style=thick, inner sep=0pt]
\tikzstyle{ent}=[circle, draw, minimum size=10mm, fill=porange]
\tikzstyle{val}=[circle, draw, minimum size=10mm, fill=plila]
\tikzstyle{tok}=[circle, draw, minimum size=10mm, fill=grau]


\node[ent] (y) {$y$} ;
\node[ent, right=of y] (z) {$z$} ;
\node[ent, left=of y] (x) {$x$} ;

\draw (y) -- (z) node[midway, above, color=black] {\textsc{arg2}} ;
\draw (y) -- (x) node[midway, above, color=black] {\textsc{arg1}} ;

\node[below right=3.5mm and 2.5mm of x, anchor=north west] {\textcolor{orange}{$\in \mathcal{X}$}} ;


\node[val, below=of x] (tx) {$t_{p,\,x}$} ;
\node[val, below=of y] (ty) {$t_{q,\,y}$} ;
\node[val, below=of z] (tz) {$t_{r,\,z}$} ;

\tikzset{style=ultra thick}
\draw[-angle 60] (x) -- (tx);
\draw[-angle 60] (y) -- (ty);
\draw[-angle 60] (z) -- (tz);

\node[below right=3.5mm and 2.5mm of tx, anchor=north west] {\textcolor{lila}{$\in \left\{\bot,\top\right\}$}} ;


\node[val, above=of x] (new) {$t_{a,\,x}$} ;
\draw[-angle 60] (x) -- (new);

\node[below right=3.5mm and 2.5mm of new, anchor=north west] {\textcolor{lila}{$\in \left\{\bot,\top\right\}$}} ;

\node[right = 9mm of new] {\LARGE?};

\end{tikzpicture}

\caption{
Inferring if $a$ is true of $x$,
when $x$ is part of a larger situation,
and given the truth value for one predicate for each pixie in the situation.
}
\label{fig:infer-context}
\end{subfigure}


\caption{
Examples of graphical models for inference.
In both cases, we want to find the probability of the predicate $a$ being true of a pixie $x$,
given some other information about the situation.
}
\label{fig:infer}


\end{figure}
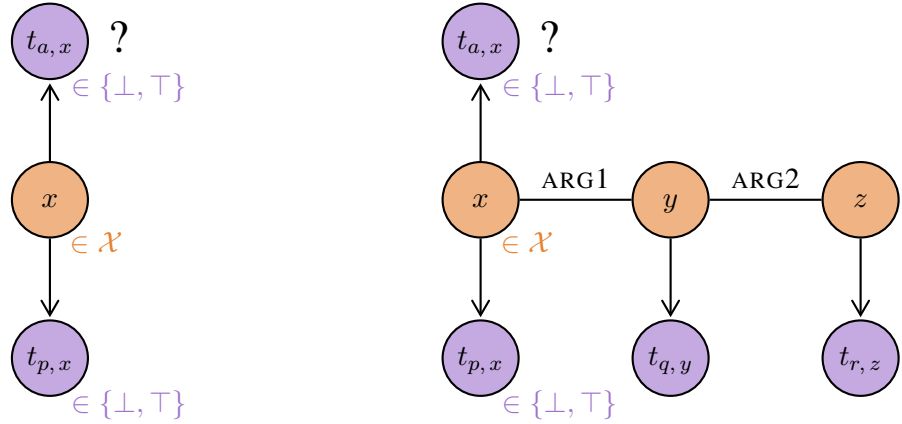

A semantic function model includes a random variable 
for the truth of each predicate for each pixie.
As noted by E\&C2,
these random variables allow us to convert certain logical propositions
into statements about conditional probabilities.
For example, we might be interested in whether one predicate implies another.
For simplicity, we can first consider a situation containing only a single pixie~$x$,
as shown in \cref{fig:infer-one}.
Then, the proposition ${\forall x\in\mathcal{X},\; p(x) \Rightarrow a(x)}$
is equivalent to the statement ${P(t_{a,x}|t_{p,x}) = 1}$.
Conditioning on $t_{p,x}$ means restricting to those pixies~$x$
for which the predicate $p$ is true,
and if the probability of $t_{a,x}$ being true is 1, then it is always true.
Similarly, ${\exists x\in\mathcal{X},\; a(x) \land b(x)}$
is equivalent to ${P(t_{a,x}|t_{p,x}) > 0}$.
This equivalence is discussed in more detail by E\&C2.

In practice, the conditional probability ${P(t_{a,x}|t_{p,x})}$ will never be exactly $0$ or $1$,
as discussed in~\cref{sec:sem-func}.
Nonetheless, this quantity
represents the degree to which $a$ implies $b$, in an intuitive sense:
the higher the value, the closer we are to \textit{every};
and the lower the value, the closer we are to \textit{no}.
We will use this quantity in~\cref{sec:lex-sim} to measure semantic similarity.

To calculate ${P(t_{a,x}|t_{p,x})}$, we need to marginalise out $x$,
because the model defines the joint probability ${P(x,\,t_{p,x},\,t_{a,x})}$.
This is analogous to the process of removing bound variables in~\cref{sec:quant},
but note that here we do not have quantifiers to evaluate.
Rather, we know certain facts about a situation,
so we want to consider just those situations where those facts are true.
Exactly marginalising out a pixie would require
summing over the entire semantic space $\mathcal{X}$,
which is intractable for a large number of dimensions.
As explained in~\cref{sec:meanfield},
the posterior for $x$ given $t_{p,x}$
can be approximated using a mean field vector.
This gives us a probability for each dimension of $x$,
representing a `typical' pixie for the observed truth values.
Applying the semantic function for $a$ to this mean field vector
lets us approximately calculate ${P(t_{a,x}|t_{p,x})}$.

In the general case, we have more than one pixie in a situation,
as shown in \cref{fig:infer-context}.
For example, if we know that a person is cutting grass,
we could ask how likely it is that the person is also
a gardener (likely), an artist (less likely), or a flowerpot (very unlikely).
As before, we can answer this question by calculating a conditional probability:
${P(t_{a,\,x}|t_{p,\,x},t_{q,\,y},t_{r,\,z})}$.
Because the truth values are connected via the latent pixies,
the truth of one predicate depends on all the others.
Inference requires marginalising out all pixies in the situation,
so we first find the joint mean field distribution for all pixies,
and then apply the semantic function for \textit{gardener}
to the mean field vector for the \textit{person} pixie.
Note how the context-dependent meaning of \textit{person}
(the mean field vector)
is crucial to this calculation --
although we are only applying applying the \textit{gardener} function
to the \textit{person} vector,
this vector depends on all predicates in the context.

\section{Experimental Results}
\label{sec:eval}

We trained our model using WikiWoods\footnote{\url{http://moin.delph-in.net/WikiWoods}},
a corpus providing DMRS graphs for 55m sentences of English (900m tokens).
WikiWoods was produced by \citet{flickinger2010wikiwoods} and \citet{solberg2012wikiwoods}
from the July 2008 dump of the full English Wikipedia,
using the English Resource Grammar \citep{flickinger2000erg,flickinger2011erg}
and the PET parser \citep{callmeier2001pet,toutanova2005pet},
with parse ranking trained on the manually treebanked subcorpus WeScience \citep{ytrestol2009wescience}.
It is distributed by DELPH-IN.

We extracted SVO triples (in a slight abuse of terminology),
by which we mean DMRS subgraphs comprising a verbal predicate
and nominal \textsc{arg1} and/or \textsc{arg2},
discarding pronouns and named entities.
This gives 10m full SVO triples,
and a further 21m where one of the two arguments is missing.
For further details, see E\&C.
To preprocess the corpus, we used the python packages
pydelphin\footnote{\url{https://github.com/delph-in/pydelphin}} (developed by Michael Goodman),
and pydmrs\footnote{\url{https://github.com/delph-in/pydmrs}} \citep{copestake2016pydmrs}.
Our source code is available online.\footnote{\url{https://github.com/guyemerson/sem-func}}

Carefully initialising the model parameters allows us to drastically reduce the necessary training time.
We initialised the parameters of the semantic functions
using random positive-only projections,
a simple random-indexing technique introduced by \citet{qasemizadeh2016vector}.
The total number of dimensions is fixed,
and each context predicate is randomly assigned to a context dimension
(which means that many contexts will be randomly assigned to the same dimension).
For each target predicate,
we count how many times each context dimension appears.
With these counts, we can calculate a standard PPMI vector.
This method lets us initialise vectors in very little time,
and we can use the same hyperparameters discussed by \citet{levy2015hyperparam}.
However, it should be noted that,
because we are not using the vectors in the same way,
the ideal hyperparameters are not the same.
In particular, we found that, unlike for normal word vectors,
it was unhelpful to use a negative offset for PPMI scores.

Once the semantic function parameters have been initialised,
the CaRBM parameters can be initialised based on mean field vectors.
Each semantic function defines a no-context mean field vector,
as described in~\cref{sec:infer} for \cref{fig:infer-one}.
For each SVO triple in the training data,
we can take the mean field vectors for the observed predicates,
and for each link, we can calculate
the mean field activation of each pair of dimensions of the linked pixies --
this is simply the outer product of the mean field vectors for the linked pixies.
We can then average these mean field activations across the whole training set,
and calculate PPMI scores,
which we can use to initialise the link's parameters.
For an average mean field activation of~$f$,
the PPMI is ${\log(f)-2\log(\frac{C}{D})}$,
where $D$ is the dimensionality and $C$ the cardinality,
since the expected activation of a pair of dimensions of two random vectors is $(\frac{C}{D})^2$.

We compare our model to two vector baselines.
The first is a standard Word2Vec model \citep{mikolov2013vector},
trained on the plain text version of the WikiWoods corpus.
The second is the same Word2Vec algorithm,
trained on the SVO triples we used to train our model:
each triple was used to produce one `sentence',
where each `token' is a predicate.

Finding a good evaluation task is far from obvious.
Simple similarity tasks do not require semantic structure,
while tasks like textual entailment
require a level of coverage beyond the scope of this paper.
We consider the SVO similarity and RELPRON datasets, described below,
because they provide restricted tasks in which we can explore approaches to semantic composition.
The results on RELPRON were also reported by E\&C2,
but we give further error analysis here.
In future work, we plan to use the datasets produced by
\citet{herbelot2016quantifier} and \citet{herbelot2013quantifier},
where pairs of `concepts' (such as \textit{tricycle}) and `features' (such as \textit{is small})
are annotated with suitable quantifiers
(out of these options: \textit{all}, \textit{most}, \textit{some}, \textit{few}, \textit{no}).
One challenge posed by these datasets is
the syntactic variation in the features,
such as \textit{has 3 wheels} and \textit{lives on coasts}.
These datasets can be seen as a further stepping stone
between this paper and general textual entailment.

\pagebreak

\subsection{Lexical Similarity}
\label{sec:lex-sim}

We evaluated our model on several lexical similarity datasets.
Our aim is firstly to show that the performance of our model
is competitive with state-of-the-art vector space models,
and secondly to show that our model can specifically target \textit{similarity} rather than \textit{relatedness}.
For example, while the predicates \textit{painter} and \textit{painting} are related,
they are true of very different individuals.

We used
SimLex-999 \citep{hill2015simlex} and SimVerb-3500 \citep{gerz2016simverb},
which both aim to measure similarity, not relatedness;
MEN \citep{bruni2014men}; and
WordSim-353 \citep{finkelstein2001wordsim}, 
which \citet{agirre2009wordsim} split into similarity and relatedness subsets.

To calculate a similarity score in our model,
we can use the conditional probability of one predicate being true,
given that another predicate is true,
as shown in \cref{fig:infer-one}.
To make this into a symmetric score,
we can multiply the conditional probabilities in both directions.
Results are shown in \cref{tab:lex-sim}.\footnote{%
  Performance of Word2Vec on SimLex-999 is higher than reported by \citet{hill2015simlex}.
  Despite correspondence with the authors,
  it is not clear why their figures are so low.
}

We can see that the semantic function model
is competitive with Word2Vec,
but has qualitatively different behaviour,
as it has very low correlation for the relatedness subset of WordSim-353.
It has lower performance on MEN and the similarity subset of WordSim-353,
but these two datasets were not annotated to target similarity, in the sense given above.
For SimLex-999 and SimVerb-3500, which do target similarity,
performance is higher than Word2Vec.

We note also that the performance of our model is higher than
that reported in our previous work.
This is due to better hyperparameter tuning.
Using the initialisation method described above
allowed for faster experiments and hence a greater exploration of the hyperparameter space.
Using more datasets also allowed for more targeted tuning:
hyperparameters for each dataset were tuned on the remaining datasets,
except for SimVerb-3500, which has its own development set.
Compared to E\&C2's results,
performance is much improved on the verb subset of SimLex-999,
which was previously tuned on noun datasets only,
indicating that the optimal settings for nouns and verbs differ considerably.

\begin{table*}[t]
\centering
\begin{tabularx}{.95\textwidth}{|l|C|C|C|C|C|C|}

\hline
Model & SL Noun & SL Verb & SimVerb & MEN & WS Sim & WS Rel \\ \hline
Word2Vec & .40 & .23 & .21 & \bf .62 & \bf .69 & .46 \\
SVO Word2Vec & .44 & .18 & .23 & .60 & .61 & .24 \\
Semantic Functions & \bf .46 & \bf .25 & \bf .26 & .52 & .60 & \bf .16 \\ \hline

\end{tabularx}
\caption{
Spearman rank correlation with average annotator judgements,
for SimLex-999 (SL) noun and verb subsets,
SimVerb-3500,
MEN, and
WordSim-353 (WS) similarity and relatedness subsets.
Note that we would like to have a \emph{low} score for WS Rel
(which measures relatedness, rather than similarity).
}
\label{tab:lex-sim}
\end{table*}

\begin{table*}[t]
\center
\begin{tabularx}{.95\textwidth}{|l|C|C|C|}

\hline
Model & GS2011 & \small RELPRON Dev & \small RELPRON Test \\ \hline
Word2Vec, Addition & .12 & .50 & .47 \\
SVO Word2Vec, Addition & .30 & -- & -- \\
Semantic Functions & .25 & .20 & .16 \\
(SVO) Word2Vec and Sem-Func Ensemble & \bf .32 & \bf .53 & \bf .49 \\ \hline

\end{tabularx}
\caption{
Spearman rank correlation with average annotator judgements,
on the GS2011 dataset, and
mean average precision on the RELPRON development and test sets.
For RELPRON, the Word2Vec model was trained on a larger training set,
so that we can directly compare with \citeauthor{rimell2016relpron}'s results.
For GS2011, the ensemble model uses SVO Word2Vec,
while for RELPRON, it uses normal Word2Vec.
}
\label{tab:beyond}
\end{table*}

\subsection{Similarity in Context}
\label{sec:cont-sim}

\citet{grefenstette2011svo} produced a dataset of pairs of SVO triples,
where only the verb varies in the pair.
Each pair was annotated for similarity.
For example, annotators had to judge the similarity of the triples
(\textit{table}, \textit{show}, \textit{result}) and
(\textit{table}, \textit{express}, \textit{result}).
In line with lexical similarity datasets,
a system can be evaluated using the Spearman rank correlation
between the system's scores and the average annotations.

For each triple, we calculated the mean field vector for the verb, conditioned on all three predicates.
We then calculated the probability that the other verb's predicate is true of this mean field vector,
similarly to \cref{fig:infer-context}
(the only difference being that we are interested in pixie $y$, not pixie $x$).
To get a symmetric score, we multiplied the probabilities in both directions.

Results are given in the ``GS2011'' column of \cref{tab:beyond}.
The performance of our model (.25) matches the best model \citeauthor{grefenstette2011svo} consider.
The performance of our ensemble (.32) matches the improved model of \citet{grefenstette2013regression},
despite using less training data.
Furthermore, the fact that the ensemble
outperforms both the semantic function model and the vector space model
shows that the two models have learnt different kinds of information.
This is not simply due to the combined model having a larger capacity --
increasing the size of the individual models did not give this improvement.

\subsection{Composition of Relative Clauses}
\label{sec:relpron}

The RELPRON dataset was produced by \citet{rimell2016relpron}.
It consists of a set of `terms', each paired with up to ten `properties'.
Each property is a short phrase,
consisting of a hypernym of the term,
modified by a relative clause with a transitive verb.
For example,
a \textit{telescope} is a \textit{device that astronomers use},
and a \textit{saw} is a \textit{device that cuts wood}.
The task is to identify the properties which apply to each term,
construed as a retrieval task:
given a single term, and the full set of properties,
the aim is to rank the properties,
with the correct properties at the top of the list.
There are $65$ terms and $518$ properties in the development set,
and $73$ terms and $569$ properties in the test set.

Since every property follows one of only two patterns
(subject or object relative clause),
this dataset lets us focus on evaluating semantics, rather than parsing.
A model that uses relatedness can perform fairly well on this dataset --
for example, \textit{astronomer} can predict \textit{telescope},
without knowing what relation there is between them.
However, the dataset also includes lexical confounders --
for example, a \textit{document that has a balance} is a financial \textit{account},
not the quality of \textit{balance} (not falling over).
The textual overlap means that
a vector addition model is easily fooled by such confounders,
and indeed the best three models that \citeauthor{rimell2016relpron} tested
all ranked this confounding property at the top.

We can represent each property as a situation of three pixies, as in \cref{fig:infer-context}.
Although they are syntactically noun phrases, the argument structure is the same as a transitive clause.
For each property, we calculated the contextual mean field vectors,
conditioned on all three predicates.
To find the probability that the term's predicate is true,
we apply the term's semantic function to the hypernym's mean-field vector.
The difference between subject and object relative clauses
is captured by whether this vector corresponds to
the \textsc{arg1} pixie or the \textsc{arg2} pixie.

Results are given in the last two columns of \cref{tab:beyond}.
Our model performs worse than vector addition,
perhaps as expected, since it does not capture relatedness,
as explained in~\cref{sec:lex-sim}.
However, the ensemble performs better than either model alone --
just as argued in~\cref{sec:cont-sim},
this shows that our model has learnt different information from the vector space model.
In particular, the ensemble improves performance on the lexical confounders.
of which there are 27 in the test set.
The vector space model places 17 of them in the top rank,
and all of them in the top 4 ranks.
The ensemble model, however, succeeds in moving 9 confounders out of the top 10 ranks.
To our knowledge, this is the first system that manages to improve both overall performance
as well as performance on the confounders.

\section{Conclusion}

We can interpret Functional Distributional Semantics as learning a probabilistic model structure,
which gives us natural operations for composition, inference, and context dependence,
with applications in both computational and formal semantics.
Our experiments show that the additional structure of the model
allows it to learn and use information that is not captured by vector space models.

\pagebreak

\section*{Acknowledgements}

We would like to thank Emily Bender,
for helpful discussion and detailed feedback on an earlier draft.
This work was supported by a Schiff Foundation studentship.

\bibliographystyle{chicago}
\bibliography{iwcs-2017}

\end{document}